\documentclass[a4paper,journal]{IEEEtran}

\usepackage{amsmath,amssymb,amsfonts,bm}
\usepackage{algorithmic}
\usepackage{algorithm}
\usepackage{array,booktabs,caption,subcaption,tabularx}
\usepackage{enumitem}
\usepackage{textcomp}
\usepackage{stfloats}
\usepackage{diagbox}
\usepackage{graphicx}
    \graphicspath{{Images/}}
\usepackage[nospace]{cite}
\usepackage{soul,verbatim}
\usepackage{tcolorbox}
\usepackage{hyperref} 
\usepackage[capitalize,nameinlink]{cleveref}
    \Crefname{figure}{Fig.}{Figs.}
    \crefname{figure}{fig.}{figs.}
    \crefname{section}{Sec.}{Secs.} 
    \Crefname{section}{Section}{Sections} 
    \Crefname{table}{Table}{Tables} 
    \crefname{table}{Tab.}{Tabs.} 
    
\newcommand{\mA}{\mathbf{A}}

\newcommand{\mC}{\mathbf{C}}

\newcommand{\mF}{\mathbf{F}}

\newcommand{\mT}{\mathbf{T}}

\newcommand{\mX}{\mathbf{X}}

\DeclareMathOperator{\diag}{diag}

\begin{document}

\title{Robust Physical Adversarial Patches\\ Using Dynamically Optimized Clusters}

\author{Harrison Bagley, Will Meakin, Simon Lucey, Yee Wei Law, Tat-Jun Chin
\thanks{This material is based upon work supported by the Air Force Office of
Scientific Research under award number FA2386-23-1-4082.}
\thanks{Bagley, Meakin, Lucey and Chin are with the Australian Institute for Machine Learning, The University of Adelaide.}
\thanks{Law is with UniSA STEM, University of South Australia.}
\thanks{Meakin and Chin have been supported by SmartSat CRC.}}

\markboth{}%
{Bagley \MakeLowercase{\textit{et al.}}: Robust Physical Adversarial Patches Using Dynamically Optimized Cluster}


\maketitle

\begin{abstract}
Physical adversarial attacks on deep learning systems is concerning due to the ease of deploying such attacks, usually by placing an adversarial patch in a scene to manipulate the outcomes of a deep learning model. Training such patches typically requires regularization that improves physical realizability (e.g., printability, smoothness) and/or robustness to real-world variability (e.g. deformations, viewing angle, noise). One type of variability that has received little attention is scale variability. When a patch is rescaled, either digitally through downsampling/upsampling or physically through changing imaging distances, interpolation-induced color mixing occurs. This smooths out pixel values, resulting in a loss of high-frequency patterns and degrading the adversarial signal. To address this, we present a novel superpixel-based regularization method that guides patch optimization to scale-resilient structures. Our approach employs the Simple Linear Iterative Clustering (SLIC) algorithm to dynamically cluster pixels in an adversarial patch during optimization. The Implicit Function Theorem is used to backpropagate gradients through SLIC to update the superpixel boundaries and color. This produces patches that maintain their structure over scale and are less susceptible to interpolation losses. Our method achieves greater performance in the digital domain, and when realized physically, these performance gains are preserved, leading to improved physical performance. Real-world performance was objectively assessed using a novel physical evaluation protocol that utilizes screens and cardboard cut-outs to systematically vary real-world conditions.
\end{abstract}

\begin{IEEEkeywords}
Physical adversarial attack, evasion attack, superpixel, Implicit Function Theorem.
\end{IEEEkeywords}

\section{Introduction}

\IEEEPARstart{C}{omputer} vision pipelines based on deep neural networks (DNNs) are susceptible to attacks that exploit these models' lack of robustness~\cite{szegedy2014intriguing}. Evasion attacks, in particular, manipulate test samples, creating adversarial examples to evade detection or cause misclassification by a trained model~\cite{nist.ai.100-2e2023}. The physical implementation of an evasion attack, called \emph{physical adversarial attack} (PAA), manipulates objects in the physical environment in which the trained model is tested on~\cite{kurakin2017adversarial}. Compared to digital adversarial attacks, PAAs are more concerning~\cite{wei2024physical} because the attacker does not need access to the digital input of the victim model; the attacker only needs to be able to conduct the physical manipulation in the scene of interest.

\begin{figure}[t]\centering
  \subfloat[\textrm{Testing adversarial patches in the real world using screens to control the position, scale and orientation of the patch, with cardboard cut-outs to control the subject's pose. See \texttt{physical\_evaluation.mp4} for demonstration.}\label{fig:cover_image}]{
    \includegraphics[width=\linewidth]{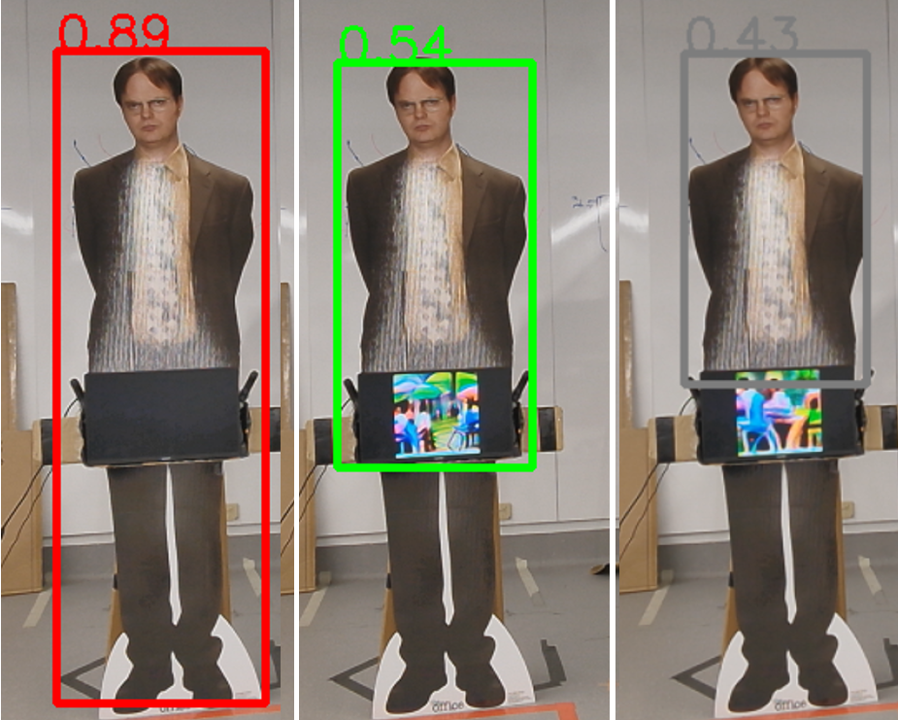}
  }
  \vspace{0.5em}

  \subfloat[\textrm{SPAP-1}\label{fig:spap-1}]{
    \includegraphics[width=0.32\linewidth]{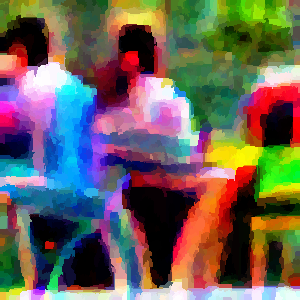}
  }\hfill
  \subfloat[\textrm{SPAP-2}\label{fig:spap-2}]{%
    \includegraphics[width=0.32\linewidth]{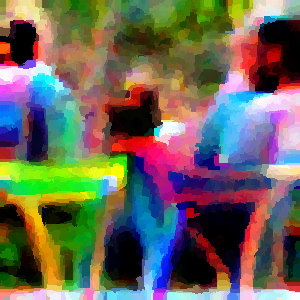}%
  }\hfill
  \subfloat[\textrm{Unclustered patch}\label{fig:unclustered}]{%
    \includegraphics[width=0.32\linewidth]{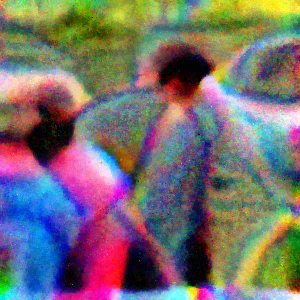}%
  }

  \caption{Testing our Superpixel Adversarial Patches (SPAPs).}
  \label{fig:testing}
\end{figure}
    
Among PAAs, the most common attack vectors are adversarial patches (``patches'' for short hereafter)~\cite{brown2017adversarial}, which attackers attach to objects in the scenes of interest. Patches have been successfully weaponized against facial recognition systems~\cite{sharif2016accessorize, liu2023unauthorized, wei2023adversarial} and person detectors~\cite{thys2019fooling, xu2020adversarial, wu2020making}. On the one hand, these attacks may be used to protect the perpetrator's privacy~\cite{liu2023unauthorized}; on the other, the efficacy of these attacks has serious public safety implications. For example, pedestrians attached with adversarial patches can become ``undetectable'' to autonomous vehicles that rely on computer vision to detect pedestrians. It is thus vital to investigate patch-based PAAs to inform the formulation of countermeasures.

The training or optimization of patches is typically conducted by minimizing an adversarial loss over a dataset. It is well-known that solely minimizing adversarial loss leads to patches that are difficult to fabricate or non-robust when viewed under real-world conditions~\cite{lu_no_2017, hoory_dynamic_2020, wei2024revisiting}. Therefore, it is vital to regularize the training to improve physical realizability (e.g., printability, smoothness) and/or robustness to real-world variability (e.g. deformations, viewing angle, noise).

An aspect of variability that has received little attention is scale variability. In the digital domain, scale variations occur when upsampling or downsampling patches to superimpose on targets, as well as in the resizing process of images during preprocessing for inference. In this physical domain, scale variations are introduced when viewing patches at different distances and viewing angles. These processes lead to a mixing of colors due to averaging or interpolation that smooths out high-frequency patterns. This invariably attenuates the adversarial signal, resulting in reduced attack efficacy.

Second, many prior works on patch-based PAAs have mainly evaluated attack efficacy in the digital domain. Assessment of the effectiveness of the attacks in the real world is usually limited and demonstrated qualitatively. A major contributing factor is the overhead in fabrication and testing the different patches in an objective, repeatable manner. 

\subsection{Contributions}
There are two main contributions in this paper:
\begin{enumerate}[leftmargin=1em]
\item \textbf{Superpixel-regularized PAA.} We introduce a superpixel PAA regularization module that induces more spatial structure in optimized patches. Our regularizer dynamically clusters pixels to train intelligent superpixel boundaries and colors that are scale-resilient; see bottom row of~\Cref{fig:testing}. Inspired by \cite{dong2020robust, yin2021scaling} which show that superpixels can be used to make attacks robust to image processing-based defenses, we show how the superpixel aglorithm SLIC~\cite{achanta2012slic} can be integrated into an end-to-end differentiable framework and backpropagated through to permit this intelligent clustering.
\item \textbf{Physical evaluation protocol.} We propose a novel evaluation protocol for PAAs, combining the usage of a digital screen and life-sized cardboard cut-outs. This factors out testing inconsistencies such as reprinting, human poses, and placement of patches, as demonstrated in \Cref{fig:cover_image}.
\end{enumerate}
We present experimental results supporting the idea that regularizing patch training with superpixel clustering leads to more effective attacks. Also, our physical evaluation protocol provides objective evidence of the transferability of our patch to the real world, as well as comprehensive and systematic benchmarking with previous patch designs that illustrate the higher attack efficacy of our method.




\subsection{Threat model}

We focus on patch-based PAAs against person detectors, which are more challenging to fool than classifiers~\cite{jia2022fooling}.

Our threat model is similar to those adopted elsewhere~\cite{sharif2016accessorize, wei2023adversarial, thys2019fooling}. The adversary's goal is to launch an evasion attack on a person detector in a surveillance system. The adversary is assumed to operate under the constraints where: 
\begin{itemize}
    \item The adversary's choice of attack vector is a patch, because patch-like objects (e.g., papers, tablets, smartphones, posters) readily blend into everyday surroundings. Note that in the our evaluation protocol (see \Cref{fig:cover_image} and \Cref{sec:eval}), a flat screen is used as the display medium for the adversarial attack, solely for the purpose of a repeatable and fair evaluation. In a practical attack setting, these patches can be validly applied to other media to improve stealth, such as on a T-shirt as described in \cite{xu2020adversarial}.


    \item The adversary is the person ``wearing'' the patch, which can be captured by the imaging system under a wide range of viewing conditions, including varying viewing angle, distance and position in an image frame.
\end{itemize}

In terms of knowledge and capabilities, the adversary operates in the white-box setting, i.e., the adversary has complete knowledge of the person detector model (called victim model) when creating and optimizing their patches against said model. However, the patch is expected to be effective against unknown person detector models, i.e., black-box evaluation.

\section{Related work}

Since emerging as a phenomenon of deep learning models, adversarial attacks have proven to be a worthwhile area of research \cite{szegedy2014intriguing, moosavi-dezfooli_deepfool_2016, goodfellow_explaining_2014, papernot_limitations_2016, carlini_towards_2017}. The threat model to the reliable and commercial deployment of DNNs is a realistic and valid concern \cite{eykholt_physicalobj_2018, thys2019fooling}. Although detection and defenses have been developed, attacks have also become more sophisticated in response to circumvent these \cite{carlini_evaluating_2019}. Hence, there is no guarantee of a model’s immunity. It is therefore imperative to understand the efficacy and resilience of attacks in real-world scenarios in order to compensate for this weakness. Here, we focus on the challenges inherent to the realization and robustness of adversarial examples in the physical world.

\subsection{Physical adversarial attacks}

From the first attacks on RGB imagery~\cite{kurakin2017adversarial}, PAAs have evolved to target other sensing modalities, including infrared~\cite{wei2023hotcold, wei2023physically}, multispectral imagery~\cite{du2022adversarial, wei2024unified}, and synthetic aperture radar automatic target recognition~\cite{zhang2024physically}.
Early attacks were mostly unconstrained noise across an image that had no conceptual counterpart to the real world \cite{szegedy2014intriguing, goodfellow_explaining_2014, moosavi-dezfooli_deepfool_2016, papernot_limitations_2016, wang_adversarial_2021, du_fast_2022}. However, they were shown to transfer to the physical world when printed and recaptured by a camera in \cite{kurakin2017adversarial}, thereby eliminating the need for an attacker to have direct access to the data fed into a model. Distinct from classification, object detection models are often utilized in the real world and are more difficult to attack. Nevertheless, they were first shown to be \textit{digitally} vulnerable in \cite{xie_adversarial_2017}, and then \textit{physically} in \cite{eykholt_physicalobj_2018} by incorporating restrictions for realistic fabrication. This restriction usually takes the form of a localized region, or ``patch", being a candidate for noise, but having less or no restriction on the amount of change in color. 

There is a sizeable body of literature on patch-based PAAs; see recent surveys~\cite{nguyen2024physical, wei2024physical}. Some of these patches are designed to be wearable \cite{xu2020adversarial, hu2021naturalistic, komkov2021advhat, wei2023hotcold, wei2023physically, guesmi2024dap, wei2024unified}, while others are not~\cite{thys2019fooling, huang2023t-sea, wei2024revisiting}. Most pipelines, including ours (see \Cref{sec:optimizing}), are similar to Thys et al.'s, but some use adversarial optimization~\cite{wei2024revisiting} instead. Regardless of the training technique, maintaining robustness under realistic environmental conditions in the real world is non-trivial, which we discuss now.

\subsection{Robustness and regularization}
Lu et al.~\cite{lu_no_2017} demonstrated that adversarial attacks do not generalize well to real-world scenarios due to the degradation inherent in the fabrication process and the deformations present in the real world (light, distance, angle, etc.). To account for the former, Sharif et al.~\cite{sharif2016accessorize} introduced two regularization terms into the objective function to minimize, namely Total Variation (TV) and Non-Printability Score (NPS). TV encourages perturbations to transition smoothly between color regions. This is desirable due to the limitations of fabrication and capture devices to accurately manage high-frequency noise. TV is now broadly used in PAAs. NPS pushes the color gamut of perturbations to be close to colors accurately representable by the fabrication device (e.g., printer, screen) and perceived by the capture device. However, real-world deformations still pose a problem.
An on-screen patch that dynamically changes depending on the viewing angle has been shown, even without using the NPS, to provide a level of robustness~\cite{hoory_dynamic_2020}. 
The choice of camera can also significantly affect attack quality \cite{wei2024revisiting}. Data augmentation has been shown to increase robustness and since Brown et al.~\cite{brown2017adversarial}, patch optimization pipelines have usually included an Expectation Over Transformation (EOT) module. In the EOT module, the transformations over which expectation is taken include randomized rotation, scaling, brightness adjustment, contrast adjustment, and noise addition. For robustness to different real-world conditions, more sophisticated transformations have been proposed, e.g., creases transformation~\cite{guesmi2024dap}, patch cut-out~\cite{huang2023t-sea}, spatial transformation using a spatial transformer~\cite{komkov2021advhat}, homography~\cite{wei2024unified}, thin-plate spline transformation~\cite{xu2020adversarial}, and generating 3D tiled textures \cite{hu_adversarial_2022, hu_physically_2023}.

Robustness to defenses is another key consideration when designing attacks. One category of defense focuses on improvements to models, such as adversarial training \cite{goodfellow_explaining_2014, madry_towards_2017} and distillation \cite{papernot_distillation_2016, papernot_extending_2017}. Although somewhat effective, these have proven to be easily circumventable due to the inherent vulnerabilities of DNNs themselves \cite{carlini_defensive_2016, ilyas_adversarial_2019, carlini_adversarial_2017, carlini_magnet_2017}. Another approach is to pre-process the input images such that adversarial noise is destroyed or at least detected \cite{guo_countering_2018, xu2018feature}. Utilizing only robust features for detection \cite{freitas_unmask_2020, zhao_apnet_2024, li_model-agnostic_2024} has also become common. These methods that focus on cleaning data before passing the data through a detector shows the need for adversarial noise to be coarse-grained and smooth to remain a viable threat. Structured regularization, using superpixels for example, offers an alternative and promising approach beyond typical handcrafted regularization terms and transformations.

\subsection{Superpixels for adversarial attacks}
Superpixels are a technique to segment an image into semantic regions in a simple yet effective manner. Superpixels can be thought of as an intelligent downsampling of an image resulting in smooth regions being homogenized and separated by hard boundaries. This is an enhancement to TV which only produces smoothed \textit{gradients} of color. Among the techniques for producing superpixels \cite{achanta2012slic, achanta_superpixels_2017, bandara_senanayaka_adaptive_2020, li_nice_2020}, we choose SLIC \cite{achanta2012slic} due to its simple 5-dimensional distance metric that combines 3 dimensions in the color space and 2 dimensions in spatial coordinates (see \Cref{sec:sp-reg}).

Utilizing superpixels in PAAs offers promising avenues for improving attack efficacy in realistic threat models. Dong et al.~\cite{dong2020robust} first utilized them to attack a classifier by generating homogenous adversarial noise in locally smooth regions with a static template, resulting in robustness to steganalysis and image processing based defenses. 
Superpixels can be leveraged for reversible adversarial examples, where only authorized parties can decrypt encrypted adversarial perturbations and recover the original data~\cite{liu2023unauthorized}.
Adversarial noise can also be made resilient to scaling by segmenting a random pixel-level patch template into a superpixel template~\cite{yin2021scaling}. Voronoi diagrams~\cite{hu_physically_2023} have alternatively been used for template segmentation.
We are the first to incorporate a dynamic template into the training loop of a patch, which we then realize physically.

Intuitively, allowing more superpixels in an attack leads to better attack effectiveness but reduced robustness \cite{dong2020robust, liu2023unauthorized}. This makes our contribution of a real-world evaluation at various restrictive levels important (see \Cref{sec:eval}), and we encourage the field to adopt this as a standard methodology.

\section{Optimizing physical adversarial attacks}\label{sec:optimizing}
\begin{figure*}[ht]
    \centering
    \includegraphics[width=\linewidth]{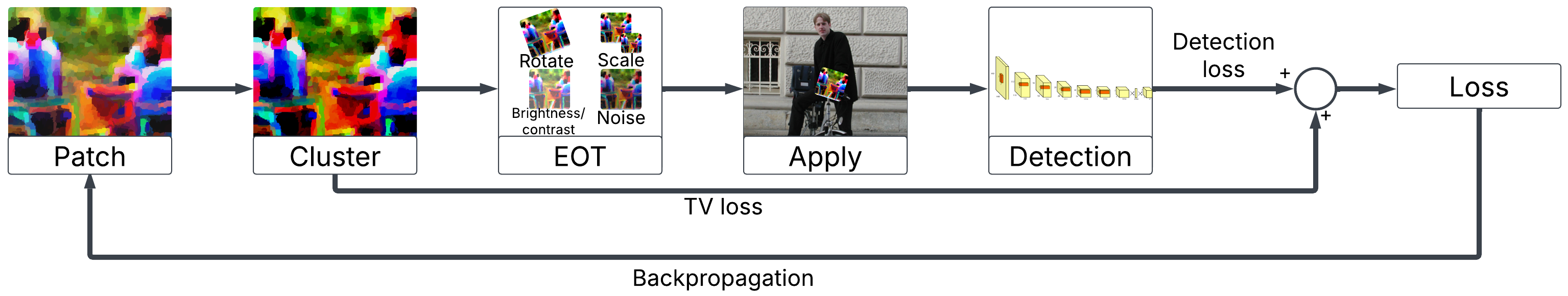}
    \captionsetup{margin=0pt,skip=0pt}
    \caption{The proposed patch training pipeline, where detection loss, TV loss and total loss are defined in Eqs.~\eqref{eqn:det_loss}--\eqref{eqn:loss}.}
    \label{fig:pipeline}
\end{figure*}

Our adversarial patch training pipeline, illustrated in \Cref{fig:pipeline}, builds on the training pipeline from Thys et al.'s~\cite{thys2019fooling}. We incorporate an additional SLIC module and a novel backpropagation mechanism (\Cref{sec:sp-reg}), to optimize our adversarial patches to allow dynamic superpixel-based regularization. The requirement for loss gradients from the object detector for backpropagation makes the proposed attack a white-box attack.

\subsection{Pipeline overview}
A patch is initialized as random noise before it is processed with SLIC and subsequently EOT, where expectation is taken over randomized rotation, scaling, brightness adjustment, contrast adjustment, and noise addition. The resultant patch is then superimposed on images of people in the INRIA person dataset~\cite{dalal2005hog}. The resultant adversarial examples are fed to a YOLOv2~\cite{redmon2017yolo9000} object detector. The choice of YOLOv2 facilitates comparison of our results with those of related works~\cite{eykholt_physicalobj_2018, thys2019fooling, hoory_dynamic_2020, wu2020making, xu2020adversarial}.

\subsection{Loss computation}
The object detector provides the detection loss, $L_{obj}$, defined as the maximum objectness score following \cite{thys2019fooling}:
\begin{equation}\label{eqn:det_loss}
    L_{obj} = \max(p_{obj,1}, p_{obj,2}, \dots, p_{obj,n}),
\end{equation}
where $p_{obj,\ast}$ is the probability that the $\ast$-th anchor point, out of $n$ anchor points, contains an object. 

We also calculate the total variation score of the patch using Eq.~\eqref{eqn:tv}, first introduced for adversarial attacks in \cite{sharif2016accessorize}:
\begin{equation}
    L_{TV} = \sum_{i,j} \sqrt{(p_{i,j}-p_{i+1,j})^2+(p_{i,j}-p_{i,j+1})^2},
    \label{eqn:tv}
\end{equation}
where $i$ and $j$ index pixels of patch $p$ in the horizontal and vertical directions respectively. The overall objective function is
\begin{equation}\label{eqn:loss}
    L = \alpha L_{TV} + L_{obj},
\end{equation}
where $\alpha=2.5$ is a scalar that has been empirically determined in \cite{thys2019fooling}. Consistent with Hoory et al.'s finding~\cite{hoory_dynamic_2020}, our use of a screen to display patches renders the NPS term unnecessary in Eq.~\eqref{eqn:loss}. The effect of NPS is nonetheless evaluated in \Cref{sec:eval}.

\subsection{Optimization}
The patch is then iteratively trained by minimizing Eq. ~\eqref{eqn:loss} with respect to the pixel values of the patch. We use the AMSGrad variant of the Adam optimizer, with an initial learning rate of 0.03 and the \textit{ReduceLROnPlateau} learning rate scheduler in minimization mode, with a patience parameter of 50 epochs.

\section{Superpixel regularization}\label{sec:sp-reg}
The previous works \cite{dong2020robust, yin2021scaling, liu2023unauthorized} apply SLIC once before their training loop to improve attack efficacy. However, we argue that this is a missed opportunity for further optimization. Inspired by Chen et al.~\cite{chen_end--end_2020}, we move the superpixel clustering stage within the training pipeline and backpropagate gradients through SLIC. A key challenge is that SLIC is typically considered non-differentiable due to the hard assignments of data to clusters. Soft assignment techniques are typically employed to circumvent the non-differentiability of the hard assignment stage \cite{hu_physically_2023, yang_superpixel_2020, jampani_superpixel_2018}. For the first time, we derive gradients for SLIC to backpropagate through, using the Implicit Function Theorem (IFT)~\cite{krantz2002implicit}, enabling \textbf{intelligent and learnable superpixel structures}.

\subsection{SLIC notation}
We begin with an $N$-pixel RGB image $\mC \in \mathbb{R}^{N \times 3}$ and aim to generate a clustered image $\hat{\mC} \in \mathbb{R}^{N \times 3}$ with $K$ superpixels and spatial sensitivity denoted by $\omega$.  

Denote by $\bm{p}_i\in\mathbb{R}^{1\times2}$ and $\bm{c}_i\in\mathbb{R}^{1\times3}$ the spatial location and RGB color for the $i$-th pixel respectively. Similarly, denote by $\bm{\lambda}_j\in\mathbb{R}^{1\times2}$ and $\bm{\phi}_j\in\mathbb{R}^{1\times3}$ the mean spatial location and mean RGB color for the $j$-th cluster respectively. Then, the following row vectors contain location and color data for the $i$-th pixel and $j$-th cluster respectively:
\begin{equation}\label{eqn:x&mu}
    \bm{x}_i \triangleq \begin{bmatrix}\bm{p}_i & \bm{c}_i\end{bmatrix}\in\mathbb{R}^{1\times5},
    \;
    \bm{\mu}_j \triangleq \begin{bmatrix}\bm{\lambda}_j & \bm{\phi}_j\end{bmatrix}\in\mathbb{R}^{1\times5}.
\end{equation}
SLIC involves optimizing the distance measure between $\bm{x}_i$ and $\bm{\mu}_j$, defined as:
\begin{equation}\label{eqn:distance}
    D(\bm{x}_i, \bm{\mu}_j) \triangleq \sqrt{\omega D_s(\bm{x}_i, \bm{\mu}_j)^2 + D_c(\bm{x}_i, \bm{\mu}_j)^2},
\end{equation}
where 
\begin{gather}
    D_s(\bm{x}_i, \bm{\mu}_j) \triangleq \|\bm{p}_i-\bm{\lambda}_j\|_2, \label{eqn:d_s} \\
    D_c(\bm{x}_i, \bm{\mu}_j) \triangleq \|\bm{c}_i-\bm{\phi}_j\|_2, \label{eqn:d_c}
\end{gather}
which measure the spatial distance and color distance respectively, between the $i$-th pixel and its cluster centroid in $L_2$ norm.

The SLIC algorithm can be formulated as minimizing a linear combination of the pixel-cluster distances:
\begin{equation}\label{eqn:obj_slic}
    \bm{\mathcal{M}}^\ast = \arg\min_{\bm{\mathcal{M}}} \sum_{i=1}^{N} \sum_{j=1}^{K} a_{ij} D(\bm{x}_i, \bm{\mu}_j),
\end{equation}
where the optimization variable is a column of $\bm{\mu}_j$'s:
\begin{equation}
    \bm{\mathcal{M}} \triangleq \begin{bmatrix}
        \bm{\mu}_1^\top & \bm{\mu}_2^\top & \cdots & \bm{\mu}_K^\top
    \end{bmatrix}^\top \in \mathbb{R}^{K \times 5}.
    \end{equation}

\subsection{Reformulating SLIC}
We define a new objective function in Eq. \eqref{eqn:obj_grad}, analogous to Eq. \eqref{eqn:obj_slic}, to permit simpler derivation of gradients:
\begin{equation}
    O(\mX, \bm{\mathcal{M}}) = \sum_{i=1}^{N} \sum_{j=1}^{K} a_{ij} \lVert \bm{\omega} \odot ( \bm{x}_i - \bm{\mu}_j ) \rVert^2_2,
    \label{eqn:obj_grad}
\end{equation}
where $\mX$ is a column of $\bm{x}_i$'s (defined in Eq.~\eqref{eqn:x&mu}):
\begin{equation}
    \mX \triangleq \begin{bmatrix}\bm{x}_1^\top & \bm{x}_2^\top & \cdots &\bm{x}_N^\top\end{bmatrix}^\top \in \mathbb{R}^{N \times 5},
\end{equation}
$\odot$ denotes the Hadamard product, and $\bm{\omega} = [\omega, \omega, 1, 1, 1]$ applies the spatial sensitivity factor to the spatial component. We also define the assignment matrix $\mA = \begin{bmatrix} a_{i,j}\end{bmatrix} \in \mathbb{R}^{N \times K}$, where $a_{i,j} = 1$ if pixel $i$ belongs to cluster $j$, or 0 if otherwise. Therefore, the clustered patch can be constructed by Eq. \eqref{eqn:cluster}:
\begin{equation}\label{eqn:cluster}
    \hat{\mC} = \mA \bm{\Phi}, 
\end{equation}
where $\bm{\Phi} \triangleq
    \begin{bmatrix}
    \bm{\phi}_1^\top & \bm{\phi}_2^\top & \dots & \bm{\phi}_K^\top
    \end{bmatrix}^\top \in \mathbb{R}^{K \times 3}$.

We consider a single dimension $d$ of $\hat{\mC}$ for $d \in \{1,2,3\}$ to obtain Eq. \eqref{eqn:assignment} by following \cite{bishop2006pattern}, holding $\mA$ constant and differentiating $\hat{\mC}$ with respect to $\bm{\Phi}_d$:
\begin{equation}\label{eqn:assignment}
    \frac{\partial \hat{\mC}_d}{\partial \bm{\Phi}_d} = \mA,
\end{equation}
where $\hat{\mC}_d \triangleq 
    \begin{bmatrix}
    \hat{\mC}_{1,d}^\top & \hat{\mC}_{2,d}^\top & \dots & \hat{\mC}_{N,d}^\top
    \end{bmatrix}^\top \in \mathbb{R}^N$.

\subsection{Applying the Implicit Function Theorem}

Define $\mF \triangleq \frac{\partial O(\mX, \bm{\mathcal{M}})}{\partial \bm{\mathcal{M}}}$ and assume $\mF$ is continuously differentiable. If 
\begin{equation}
    \mF(\mX^\ast, \bm{\mathcal{M}}^\ast) = \bm{0}
\end{equation}
and $\left[\frac{\partial \mF}{\partial \bm{\mathcal{M}}}(\mX^\ast, \bm{\mathcal{M}}^\ast) \right]^{-1}$ exists, then by the IFT~\cite{krantz2002implicit}, there exists an open set $U\in\mathbb{R}^{N\times5}$, such that $\mX^\ast \in U$, and furthermore, there exists a continuously differentiable function $g:\mathbb{R}^{N\times5}\to\mathbb{R}^{K\times5}$, such that $\bm{\mathcal{M}}^\ast = g(\mX^\ast)$ and $\mF(\mX, g(\mX)) = 0$, $\forall \mX \in U$. In other words, $g$ is a solver for $\mF = \bm{0}$, $\forall \mX \in U$. In our case, $g$ is parameterized by the number of superpixels, $K$, and the spatial sensitivity $\omega$. Moreover, by the IFT, $\forall \mX \in U$,
\begin{equation}\label{eqn:dM/dX}
    \frac{\partial g}{\partial \mX} = \frac{\partial \bm{\mathcal{M}}}{\partial \mX} = -\left[\frac{\partial \mF}{\partial \bm{\mathcal{M}}}\right]^{-1} \left[\frac{\partial \mF}{\partial \mX}\right].
\end{equation}
Eq.~\eqref{eqn:dM/dX} provides a means of computing the gradients $\frac{\partial \hat{\mC_d}}{\partial \mC_d}$, once the right-hand side of Eq.~\eqref{eqn:dM/dX} is worked out.

We first derive an expression for $\mF = \frac{\partial O(\mX, \bm{\mathcal{M}})}{\partial \bm{\mathcal{M}}} = \begin{bmatrix}\bm{f}_1^\top & \bm{f}_2^\top & \dots & \bm{f}_K^\top\end{bmatrix}^\top \in \mathbb{R}^{K\times 5}$, recalling $\bm{\mathcal{M}}$ is a column of $\bm{\mu}_j$'s:
\begin{equation}\label{eqn:F_j}
    \bm{f}_j = \frac{\partial O(\mX, \bm{\mathcal{M}})}{\partial \bm{\mu}_j} = -2 \sum_{i=1}^{N} a_{ij} \bm{\omega} \odot \bm{\omega} (\bm{x}_i - \bm{\mu}_j).
\end{equation}

To obtain $\frac{\partial \mF}{\partial \bm{\mathcal{M}}}$ in Eq.~\eqref{eqn:dM/dX}, $\bm{f}_j$ needs to be differentiated with respect to $\bm{\mu}_j$.

We again follow \cite{bishop2006pattern} and hold $a_{i,j}$ constant, assuming that a small change in pixel value will not result in a change in cluster assignment. Considering the $q$-th element of $\bm{f}_j$ for $q \in \{1, \dots, 5\}$ and partially differentiating Eq. \eqref{eqn:F_j} with respect to $\mu_{l,e}$ for $l \in \{1, \dots, K\}$ and $e \in \{1, \dots, 5\}$, we get
\begin{equation}
    \frac{\partial f_{j,q}}{\partial \mu_{l,e}} =
    \begin{cases} 
        2 \lvert \mathcal{S}_j \rvert \omega_q^2, \quad & \text{if } j = l \text{ and } q = e, \\
        0 & \text{else},
    \end{cases}
\end{equation}
where $\lvert \mathcal{S}_j \rvert$ denotes the number of elements in cluster $j$. Therefore, we have $\frac{\partial \mF}{\partial \bm{\mathcal{M}}} \in \mathbb{R}^{(K\times 5)\times (K\times 5)}$, which is a diagonal matrix and can be expressed as:
\begin{equation}
    \frac{\partial \mF}{\partial \bm{\mathcal{M}}} = 2\begin{bmatrix} \lvert \mathcal{S}_1 \rvert & 0 & \dots & 0 \\
    0 & \lvert \mathcal{S}_2 \rvert & \dots & 0 \\
    \vdots & \vdots & \ddots & \vdots \\
    0 & 0 & \dots & \lvert \mathcal{S}_K \rvert \end{bmatrix} \otimes \diag(\bm{\omega} \odot \bm{\omega}),
    \label{eqn:df/dmu}
\end{equation}
where $\otimes$ denotes the Kronecker product.

To obtain $\frac{\partial \mF}{\partial \mX}$ in Eq.~\eqref{eqn:dM/dX}, $\bm{f}_j$ needs to be differentiated with respect to $\bm{x}_i$.

Partial differentiation of Eq.~\eqref{eqn:F_j} with respect to $\bm{x}_i$ yields
\begin{equation}
    \frac{\partial \bm{f}_j}{\partial \bm{x}_i} = -2 a_{i,j} \diag(\bm{\omega} \odot \bm{\omega}).
\end{equation}
Therefore, we have $\frac{\partial \mF}{\partial \mX} \in \mathbb{R}^{(K\times 5)\times(N\times 5)}$, which can be expressed by Eq.~\eqref{eqn:df/dx}:
\begin{equation}
    \frac{\partial \mF}{\partial \mX} = -2 \mA^\top \otimes \diag(\bm{\omega}\odot\bm{\omega}).
    \label{eqn:df/dx}
\end{equation}
%
Substituting Eq.~\eqref{eqn:df/dmu} and Eq.~\eqref{eqn:df/dx} into Eq.~\eqref{eqn:dM/dX}, we finally get
\begin{equation}\begin{split}
    \frac{\partial \bm{\mathcal{M}}}{\partial \mX}
    &= \begin{bmatrix} \frac{1}{\lvert \mathcal{S}_1 \rvert} & 0 & \dots & 0 \\
    0 & \frac{1}{\lvert \mathcal{S}_2 \rvert} & \dots & 0 \\
    \vdots & \vdots & \ddots & \vdots \\
    0 & 0 & \dots & \frac{1}{\lvert \mathcal{S}_K \rvert} \end{bmatrix} \mA^\top \otimes \mathbb{I}_{5} \\
    &= \bm{\Gamma} \mA^\top \otimes \mathbb{I}_{5},
\end{split}\end{equation}
where $\bm{\Gamma}$ is a diagonal matrix with the elements $\frac{1}{\lvert \mathcal{S}_j \rvert}$ along the diagonal and $\mathbb{I}$ is the identity matrix.

When clustering the image, we no longer worry about the spatial data. Instead, we only consider the color elements. When considering the $d$-th element of the color components, we achieve Eq.~\eqref{eqn:IFT}:
\begin{equation}
    \frac{\partial \bm{\Phi}_d}{\partial \mC_d} = \bm{\Gamma} \mA^\top.
    \label{eqn:IFT}
\end{equation}
Using the chain rule and combining the results from Eq.~\eqref{eqn:assignment} and Eq.~\eqref{eqn:IFT},
\begin{equation}
        \frac{\partial \hat{\mC_d}}{\partial \mC_d} = \frac{\partial \hat{\mC_d}}{\partial \bm{\Phi}_d} \frac{\partial \bm{\Phi}_d}{\partial \mC_d}
        = \mA\bm{\Gamma}\mA^\top.
    \label{eqn:grad}
\end{equation}
Eq.~\eqref{eqn:grad} gives us the gradients that describe how each clustered image pixel changes with respect to the original image pixels during the SLIC process. Combining these gradients along with our loss function gradients using the chain rule permits backpropagation through SLIC when training our adversarial patches to permit intelligent clustering.

\subsection{Verifying gradients}
To verify our gradients, we design an alternative optimization task, unrelated to adversarial patch training. We begin with an image, $\mC \in \mathbb{R}^{N\times 3}$ and a clustered target image $\mT \in \mathbb{R}^{N\times3}$. We apply SLIC to $\mC$ to obtain a clustered image $\hat{\mC} \in \mathbb{R}^{N\times3}$. Our goal is to modify the pixel values of $\mC$ through SLIC such that its clustered representation $\hat{\mC}$ looks similar to our target $\mT$. We measure this similarity using mean squared error in Eq.~\eqref{eqn:mse}.

\begin{equation}
    L_{pixel} = \frac{1}{N}\lVert \hat{\mC}-\mT \rVert_2^2.
    \label{eqn:mse}
\end{equation}

Applying the chain rule with our gradients from Eq.~\eqref{eqn:grad} yields:
\begin{equation}
    \frac{\partial L_{pixel}}{\partial \mC} = \frac{\partial L_{pixel}}{\partial \hat{\mC}} \frac{\partial \hat{\mC}}{\partial \mC}
    = \frac{2}{N}(\hat{\mC}-\mT)^\top\mA\bm{\Gamma}\mA^\top.
    \label{eqn:toy_chain}
\end{equation}

\begin{figure*}[ht]
    \centering
    \includegraphics[width=\linewidth]{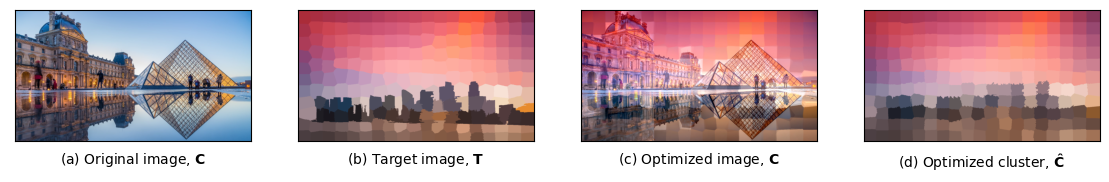}   
    \captionsetup{margin=0pt,skip=0pt}
    \caption{Backpropagating through SLIC. (a) shows our original image,  (b) shows our target image. (c) shows our image after training, (d) shows our clustered image after training. We optimize the pixel values of $\mC$ such that when we cluster it, $\hat{\mC}$ appears visually similar to our target image $\mT$. See \texttt{slic\_gradients.mp4} for visualization.}
    \label{fig:clustering}
\end{figure*}

\begin{figure}[ht]
    \centering
    \includegraphics[width=0.8\linewidth]{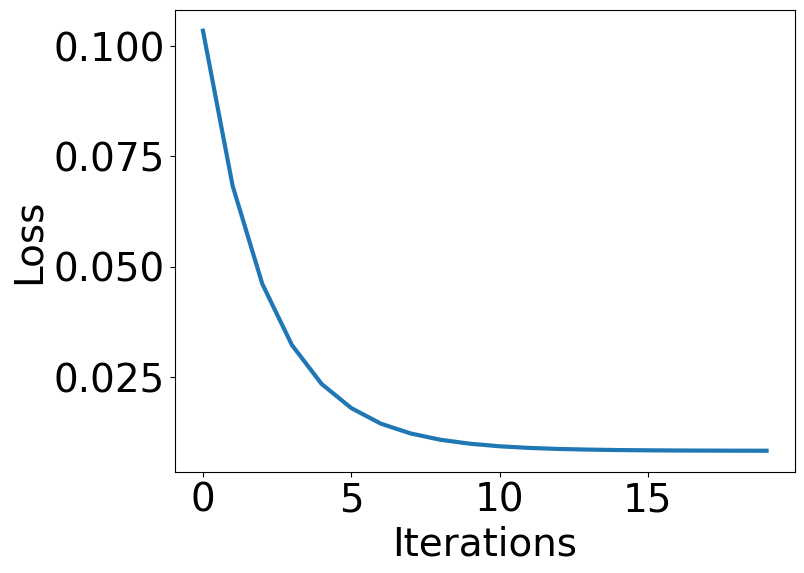}
    \caption{Training loss when backpropagating through SLIC, verifying the ability of our SLIC gradients to minimize a loss function.}
    \label{fig:loss}
\end{figure}

\Cref{fig:clustering} demonstrates the successful outcome of our training process. \Cref{fig:clustering}(d) shows that our clustered image closely resembles our target in \Cref{fig:clustering}(b). Interestingly, \Cref{fig:clustering}(c) shows that after training, artifacts from the original image remain intact, which vanish after clustering. \Cref{fig:loss} shows the convergence of our training loss with these gradients, verifying the ability to minimize $L_{pixel}$ in Eq.~\eqref{eqn:mse}. 

\section{Physical evaluation protocol}\label{sec:eval}

PAA evaluation is often conducted unsystematically in the physical domain. Most physical assessment of attacks often conduct a qualitative evaluation~\cite{thys2019fooling} or use an attack success rate (ASR) metric for video frames that does not permit a fair comparison of adversarial patches~\cite{eykholt_physicalobj_2018, hu_adversarial_2022, guesmi2024dap, xu2020adversarial}. This is problematic since physical-domain assessment is a crucial aspect of this research area to understand real-world viability, which is often inadequately assessed. Existing studies often present isolated success cases which do not reveal the extent of the limitations of patch performance in the real world. Additionally, authors often do not utilize an objective measure that permit fair comparison of patch performance.

In designing a controlled and systematic approach to physical evaluation, we consider the factors that influence patch performance. These variables include the target's pose, clothing, the scene, lighting conditions, as well as the patch's location, scale and orientation~\cite{shack2024breaking}. These factors limit fair comparison due to the difficulty of controlling all of these variables simultaneously, in addition to the large overhead of conducting physical experiments in terms of time and cost.
Evaluations are often performed digitally rather than physically, preventing a true understanding of PAA performance in a practical setting that accounts for the degradations in attack performance associated with physical realizations of adversarial patches.

\begin{figure}[t]\centering
\centering
\includegraphics[width=0.9\linewidth]{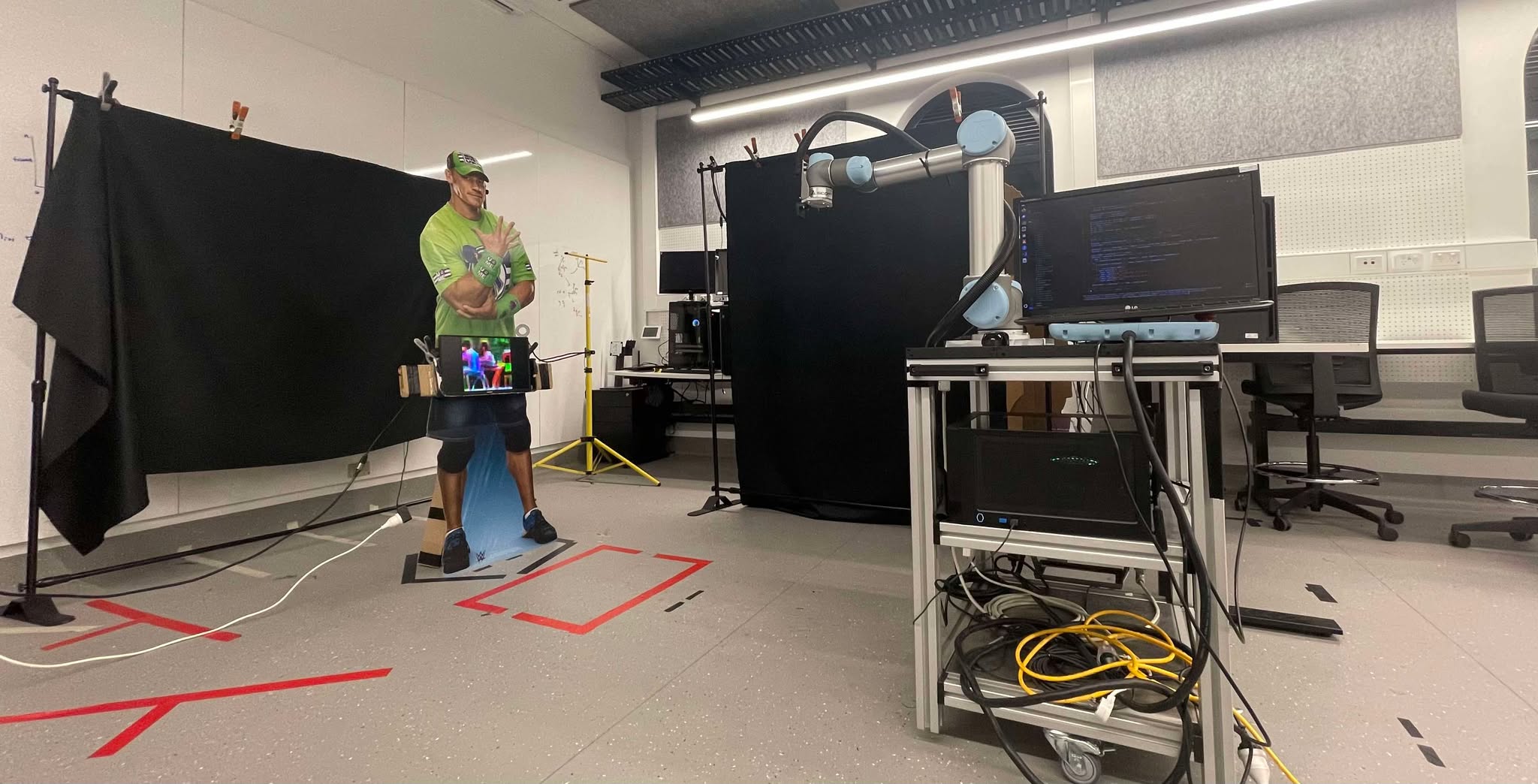}
\caption{Side view of our physical evaluation protocol setup.}
\label{fig:exp_setup}
\end{figure}

Studies such as \cite{chahe_dynamic_2024, hoory_dynamic_2020} utilize screens to explore dynamic patches. These screens permit different patches to be displayed in the exact same orientation, position and scale. Inspired by these studies, we introduce a novel evaluation protocol and apply this process to our patches. We employ life-sized cardboard cut-outs as our targets to evade detection of, and a digital screen to display our patches in a lighting-controlled laboratory. Using our setup, we are able to fix all of the aforementioned variables and display a series of patches on the screen with minimal manual effort; see \Cref{fig:exp_setup}.

We record our patched subjects and extract a number of frames that are run through the victim models. We compute the maximum detection score for each frame and find the average score for each patch. We also perform baseline tests using a blank (black) screen and a patch displaying random noise to compare the true effect of our patches in degrading detection capabilities. This allows us to produce an objective metric that is comparable across the range of patches for the given setup.

We repeat this process across a range of distances, $2\,\mathrm{m}$--$9\,\mathrm{m}$, and viewing angles from $-45^\circ$ to $+45^\circ$ side-to-side, and from $-20^\circ$ to $+20^\circ$ vertically. Tests are conducted over a range of target subjects in the form of celebrity cut-outs, namely Pitbull, John Cena, Danny Devito, Dwight Schrute, Taylor Swift and Will Ferrell.

In line with \cite{hoory_dynamic_2020}, we drop the NPS loss term when optimizing our patches. We also generate a patch using Thys et al.'s~\cite{thys2019fooling} approach without NPS, as a control.

\section{Experimentation and results}\label{sec:results}
We perform a threefold evaluation of our patches: digitally, using our physical cut-out protocol from \Cref{sec:eval}, as well as using real people in more challenging scenarios. For experiments using real people, informed consent has been obtained. We utilize the AP drop metric to assess performance digitally, and the maximum objectness/confidence scores for physical testing.

\subsection{Digital evaluation}\label{sec:digital}
\begin{figure}[h]\centering
  \begin{subfigure}{\linewidth}
    \centering
    \includegraphics[width=0.9\linewidth]{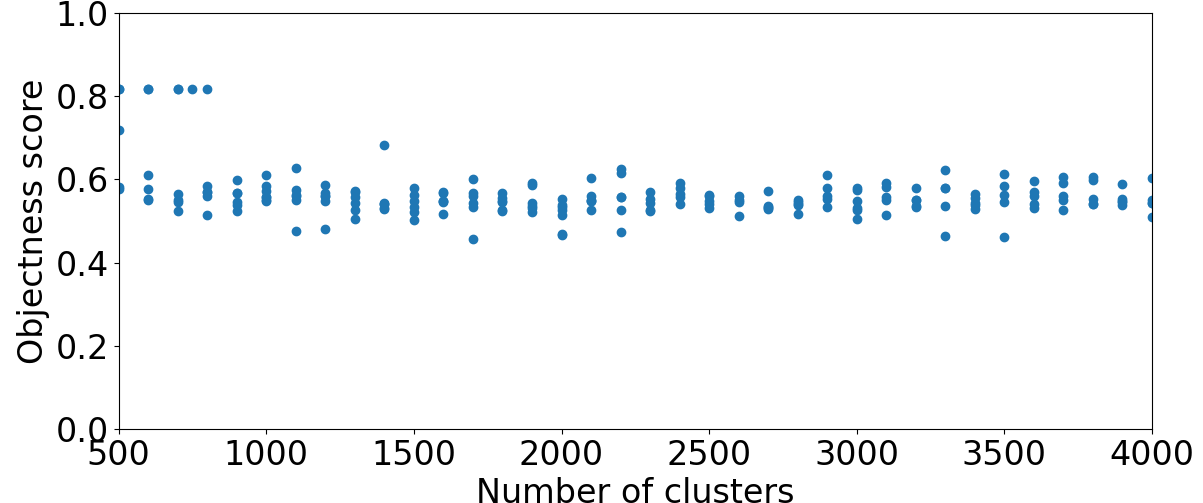}
    \caption{}
    \label{fig:digital_comparison:obj}
  \end{subfigure}  
  \begin{subfigure}{\linewidth}
    \centering
    \includegraphics[width=0.8\linewidth]{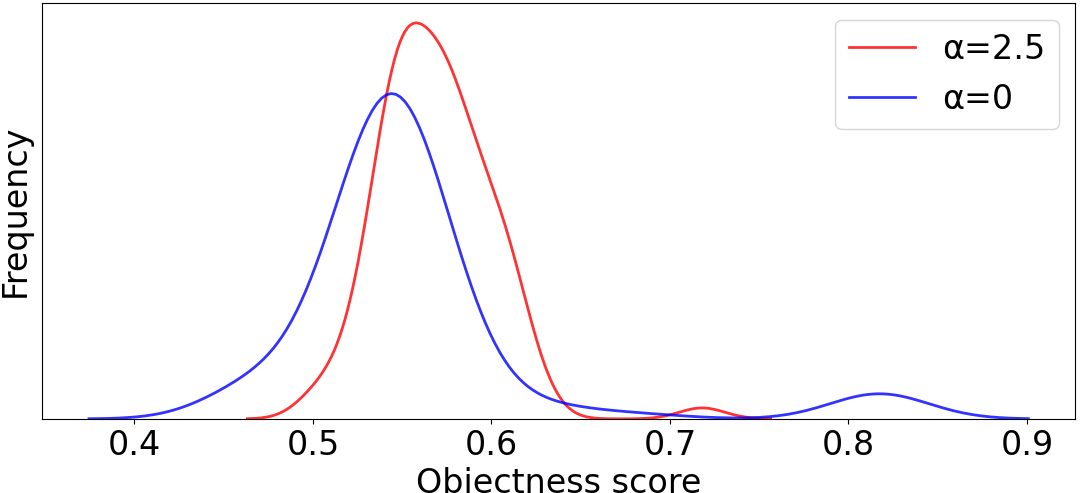}
    \caption{}    
    \label{fig:digital_comparison:tv}
  \end{subfigure}  
  \begin{subfigure}{\linewidth}
    \centering
    \includegraphics[width=0.8\linewidth]{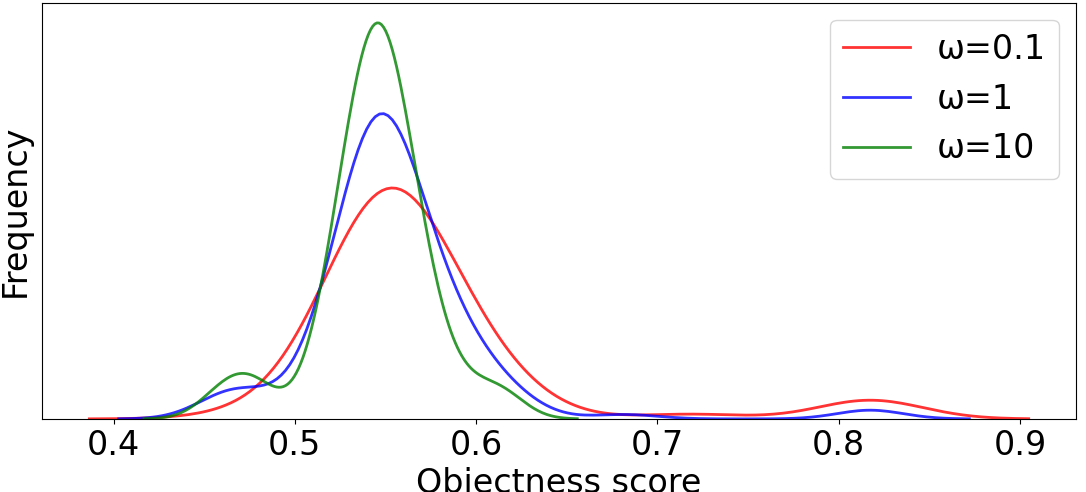}
    \caption{}
    \label{fig:digital_comparison:omega}
  \end{subfigure}
  \captionsetup{margin=0pt,skip=0pt}
  \caption{(a) Objectness score compared to the number of superpixels/clusters. (b) Distribution of patch objectness score for patches with ($\alpha=2.5$) and without ($\alpha=0$) TV loss. (c) Distribution of objectness score for different spatial sensitivities of the superpixels.}
  \label{fig:digital_comparison}
\end{figure}

The initial experiments assess the effect of tuning SLIC parameters on patch performance in the digital domain.
We generate a range of patches with $K$ superpixels where $K \in \{500, 600, \dots, 3900, 4000\}$ and vary the spatial sensitivity $\omega \in \{0.1, 1, 10\}$. We also explore training patches with and without the TV loss by setting $\alpha \in \{0, 2.5\}$. When generating our patches, we use a fixed random seed. \Cref{fig:digital_comparison:obj} shows a graph of objectness score against the number of superpixel clusters, $K$. For $K<1000$, attack performance is not effective, supported by results from \cite{dong2020robust}. At around $K=1000$, there is a step rise in attack performance (drop in objectness score) that does not improve further with larger values of $K$. 

\Cref{fig:digital_comparison:tv} shows the distribution of patch effectiveness in terms of objectness score, both with and without the TV loss in Eq.~\eqref{eqn:loss}. Both graphs have a similar mean score, indicating that there is negligible difference in attack performance in the digital domain. 

\Cref{fig:digital_comparison:omega} shows the distribution of patch effectiveness in terms of objectness score, when varying the spatial sensitivity of the superpixel clusters. The mean objectness score for each plot is approximately the same, indicating that the spatial sensitivity does not have a significant effect on attack performance.

\subsubsection{White-box evaluation}
Two categories of Superpixel Adversarial Patches are defined: \textit{SPAP-1} ($K=4000$, $\omega=0.1$, $\alpha=2.5$) and \textit{SPAP-2} ($K=3600$, $\omega=0.1$, $\alpha=2.5$).
We compare the effects of \textit{SPAP-1} and \textit{SPAP-2} to the effects of a clean sample, a random-pixel patch, \textit{AdvPatch} from \cite{thys2019fooling}, AdvPatch without NPS (\textit{AdvPatch2}), and a post-training clustered patch.
Using the test set of the INRIA person dataset~\cite{dalal2005hog}, we measure the AP drop caused by the patches.

\begin{table}[]
\centering
\begin{tabular}{@{}c|cc@{}}
    \toprule
    Patch type& \textit{AP (\%)}   \\ \midrule
    Clean     & 100   \\
    Random    & 92.23 \\
    AdvPatch  & 24.97 \\
    AdvPatch2 & 44.60 \\
    Post-clustering & 30.64 \\ 
    SPAP-1    & 22.23 \\
    SPAP-2    & 16.28 \\
    \bottomrule
    \end{tabular}
    \caption{AP scores of various patches under digital evaluation on the INRIA person test set.}
    \label{tab:digital}
\end{table}

\Cref{tab:digital} shows the results of our digital evaluation. The clean dataset acts as a baseline with an expected AP of 100\% as the ground truth labels are extracted from the victim model's predictions. We then compare the adversarial affect of a randomly generated patch as a control, which demonstrates minimal adversarial affect with an AP of 92.23\%. We show that our superpixel patches, \textit{SPAP-1} (22.23\%) and \textit{SPAP-2} (16.28\%) outperform the state-of-the-art AdvPatch (24.97\%) as evidenced by the lower \textit{AP} scores. We observe no adversarial benefit is attained by applying clustering post-training. We also observe that a lack of NPS regularization additionally does not aid adversarial performance.

\subsubsection{Black-box evaluation}\label{sec:black-box}
We further analyze our patches in a black-box setting by applying the YOLOv2 trained patches to victim models that are assumed to be unknown during patch generation. These models are chosen to be: 
\begin{itemize}
    \item the YOLOv5 series~\cite{yolov5} of single-shot detection models;
    \item Faster R-CNN~\cite{ren2017faster}, a two-shot detector; and
    \item DEtection TRansformer (DETR)~\cite{carion2020end}, a vision transformer model.
\end{itemize}
All these models have previously been used to evaluate adversarial attacks~\cite{jia2022fooling, xu2020adversarial, wei2023hotcold}.

\begin{table*}[ht]
\centering
\begin{tabular}{cccllllll}
\hline
\multicolumn{1}{c|}{\diagbox[]{Patch}{Victim model}}  & YOLOv2  & YOLOv5n        & \multicolumn{1}{c}{YOLOv5s} & YOLOv5m        & YOLOv5l        & YOLOv5x        & Faster R-CNN    & DETR           \\ \hline
\multicolumn{1}{c|}{Clean}    & 100            & 100            & 100            & 100            & 100            & 100            & 100            & 100            \\
\multicolumn{1}{c|}{AdvPatch} & 24.97          & 34.55          & \textbf{33.25} & 74.14          & 75.29          & 80.65          & 72.55          & 51.22          \\
\multicolumn{1}{c|}{SPAP-1}   & 22.23          & \textbf{27.47} & 37.73          & \textbf{62.36} & \textbf{67.70} & 78.26          & \textbf{68.86} & \textbf{48.89} \\
\multicolumn{1}{c|}{SPAP-2}   & \textbf{16.28} & 33.91          & 44.20          & 66.48          & 69.14          & \textbf{77.83} & 71.11          & 50.61          \\
\multicolumn{1}{c|}{Random}   & 92.23          & 91.89          & 93.75          & 96.69          & 86.45          & 88.40          & 94.54          & 77.68          \\ \hline
\end{tabular}
\caption{AP scores for evaluating the transferability of PAAs against various victim models. The YOLOv2 column represents the white-box attack case, while the other columns are black-box cases.}
\label{tab:digital_black}
\end{table*}

\Cref{tab:digital_black} shows the AP scores achieved when applying the different patches to the various victim models. The most effective attack for each model is highlighted in bold. We show that our superpixel adversarial attacks tend to transfer better across the black-box victim models, as evidenced by the lower AP scores compared to those for AdvPatch. The digital evaluation indicates that our attacks are strongest against YOLOv2 (white box), transfer well to YOLOv5n and YOLOv5s, and transfer moderately well to DETR.
Our attacks transfer least well to Faster R-CNN and larger YOLOv5 variants, but still outperform other attacks.

\begin{figure*}[ht]
    \centering
    \includegraphics[width=\linewidth]{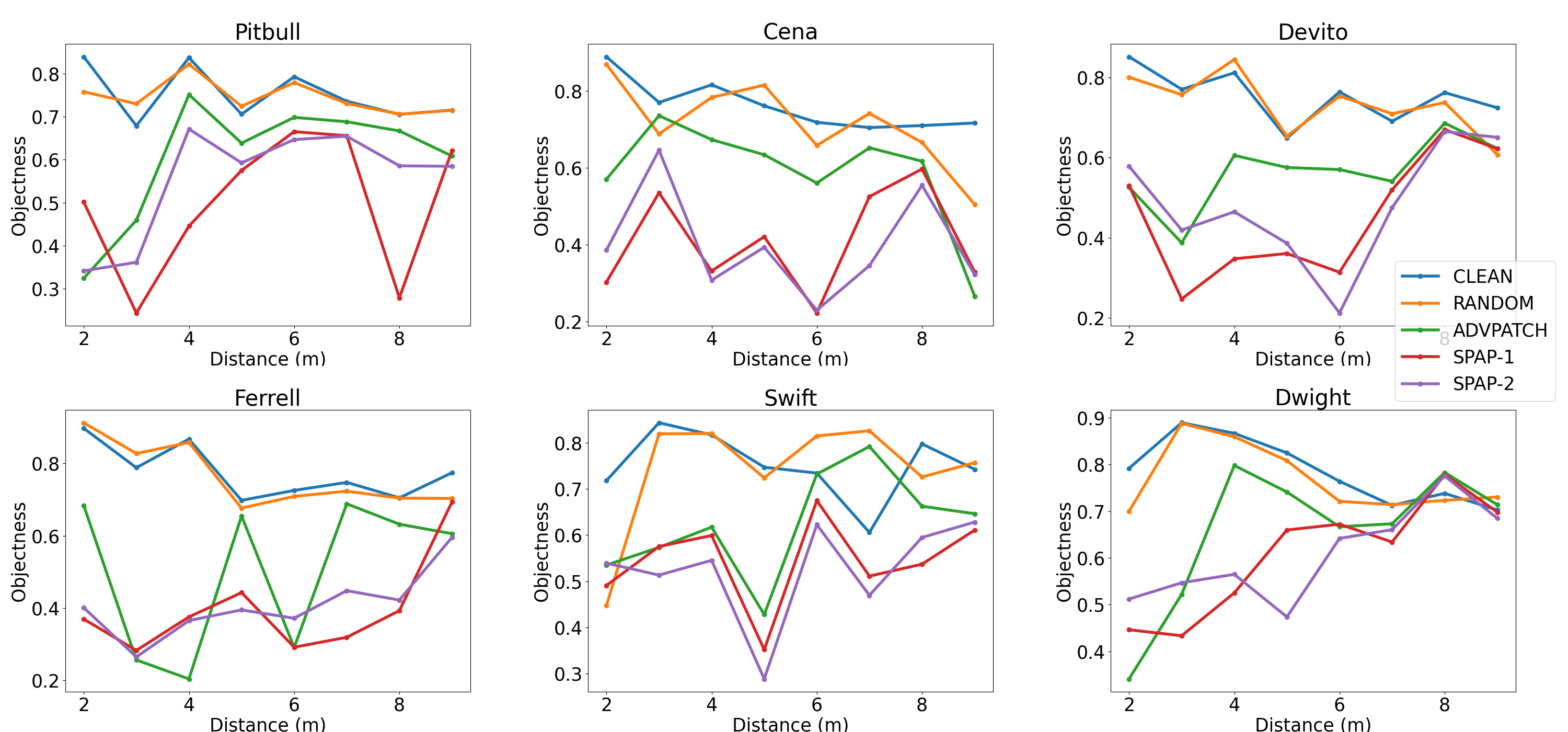}
    \caption{Physical evaluation of our adversarial patches on different cardboard cut-out subjects over distances of $2\,\mathrm{m}$--$9\,\mathrm{m}$. We use the objectness metric to convey the strength of the attack compared the clean and random patches.}
    \label{fig:distance_results}
\end{figure*}

\begin{figure*}[ht]
    \centering
    \includegraphics[width=\linewidth]{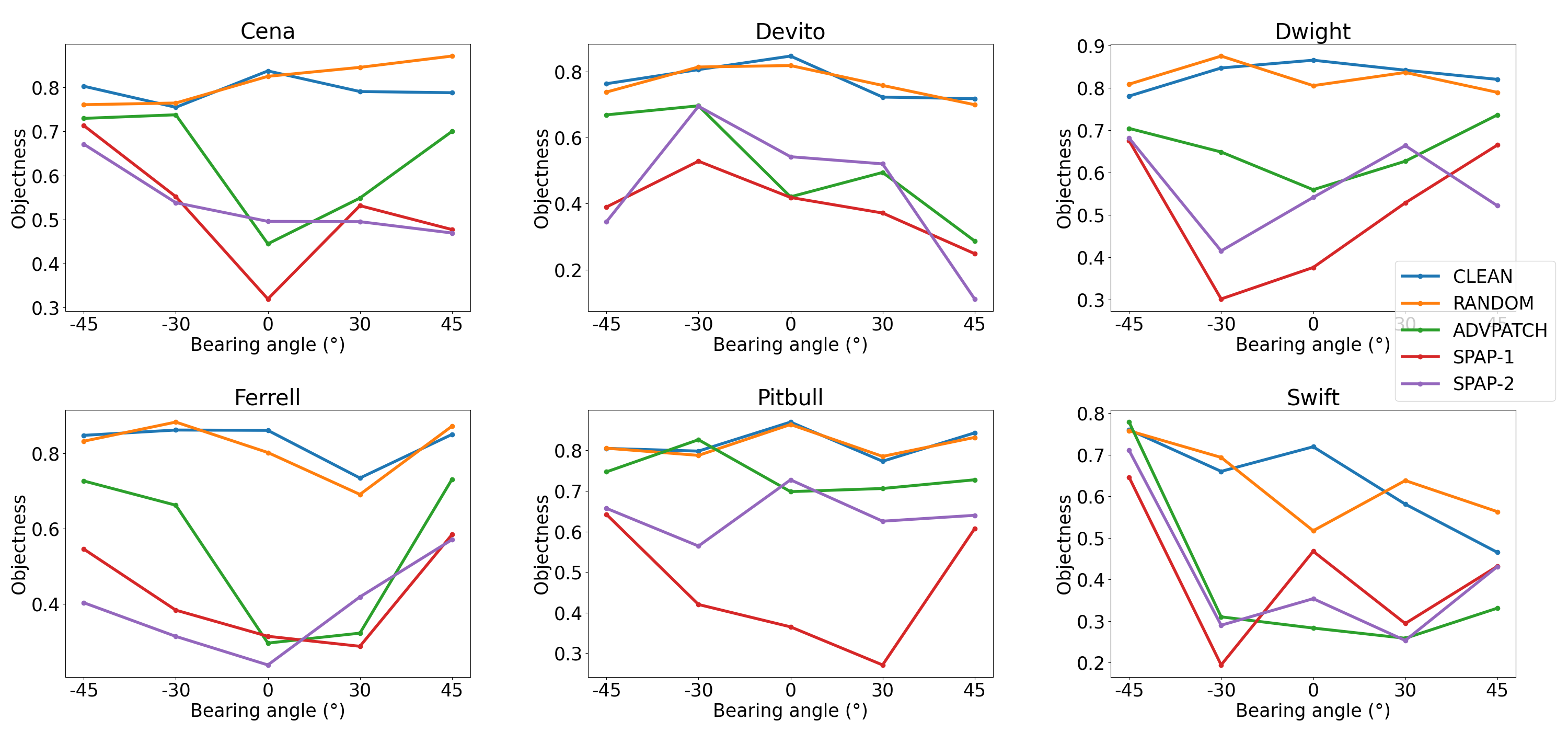}
    \caption{Physical evaluation of our adversarial patches (side-to-side variation).}
    \label{fig:side_angle_results}
\end{figure*}

\begin{figure*}[ht]
    \centering
    \includegraphics[width=\linewidth]{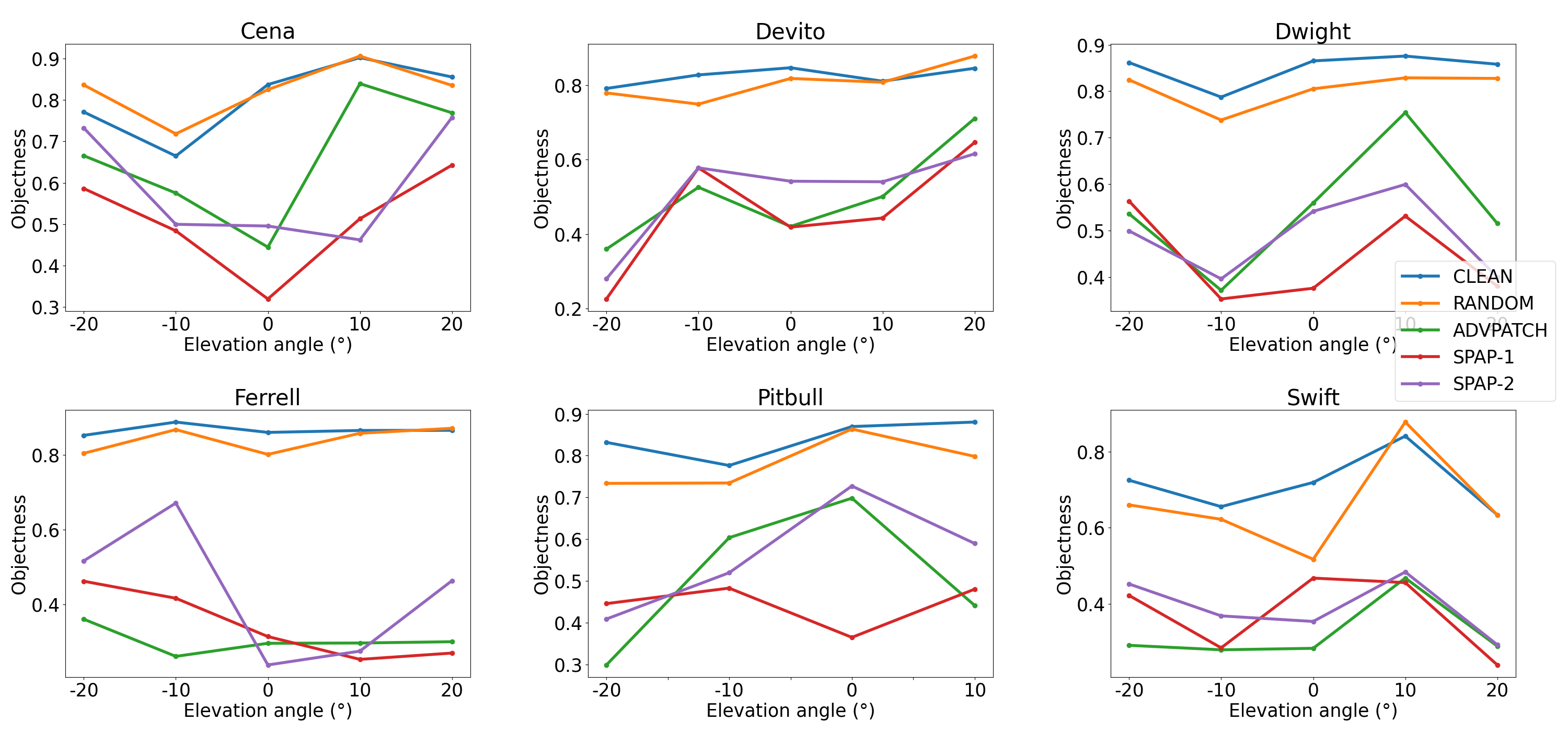}
    \caption{Physical evaluation of our adversarial patches (top-to-bottom variation).}
    \label{fig:top_angle_results}
\end{figure*}

\subsection{Physical evaluation with cut-out protocol}\label{sec:physical-cutout}
In this section, we compare the performance of various patches using our physical evaluation protocol described in \Cref{sec:eval}. We initially compare our patch performance across a range of distances and viewing angles.                        
In \Cref{fig:distance_results}, we measure the objectness score of various patches on our cut-outs at a range of distances. The results show that our patches \textit{SPAP-1} and \textit{SPAP-2} tend to outperform \textit{AdvPatch} across most of the tested distances, demonstrating better physical-domain robustness than AdvPatch. 

A general trend for each cut-out as the camera moves further away from the cut-out, is the decreasing person detectability in the absence of a patch or in the presence of a random patch. Additionally, the adversarial effect of all trained patches also tends to diminish. However, there are a number of cases where attack performance does not follow this trend, exhibiting large spikes. This observation suggests that PAA performance tends to be erratic and there are other factors that impact attack performance in the physical world, such as losses that occur during the imaging process and ambient lighting.

\Cref{fig:side_angle_results} shows patch performance across a range of bearing angles (horizontal viewing angles), and \Cref{fig:top_angle_results} shows patch performance across a range of elevation angles (vertical viewing angles). Again, in most of the tested viewing angles, our superpixel adversarial patches tend to exhibit a larger adversarial effect than AdvPatch.

\begin{figure}[ht]
    \centering
    \begin{subfigure}{0.32\linewidth}
        \centering
        \includegraphics[width=\linewidth]{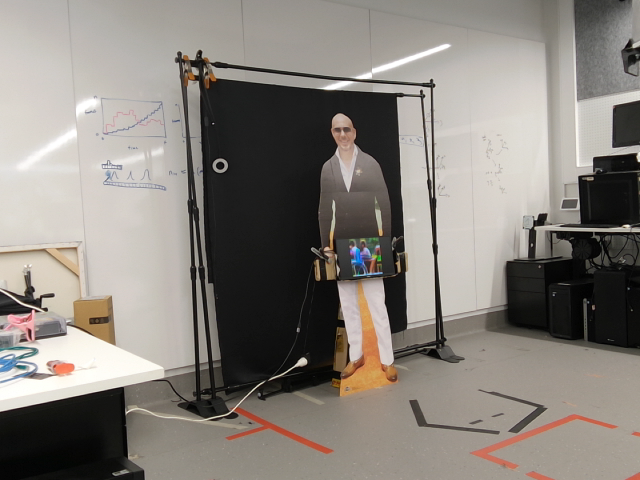}
        \caption{Side view}
        \label{fig:side}
    \end{subfigure}
    \begin{subfigure}{0.32\linewidth}
        \centering
        \includegraphics[width=\linewidth]{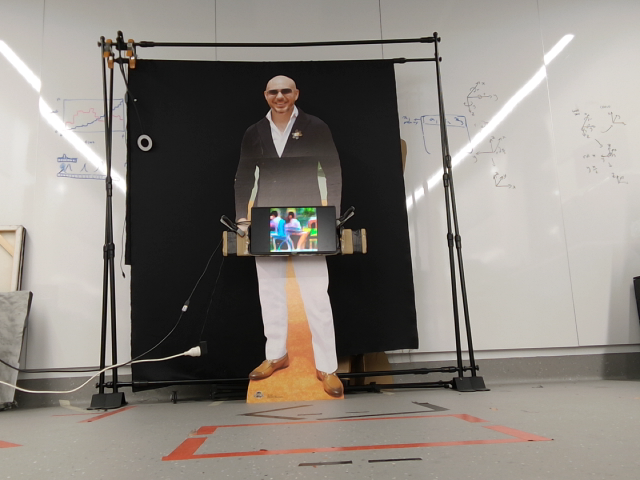}
        \caption{Bottom view}
        \label{fig:bottom}
    \end{subfigure}
    \begin{subfigure}{0.32\linewidth}
        \centering
        \includegraphics[width=\linewidth]{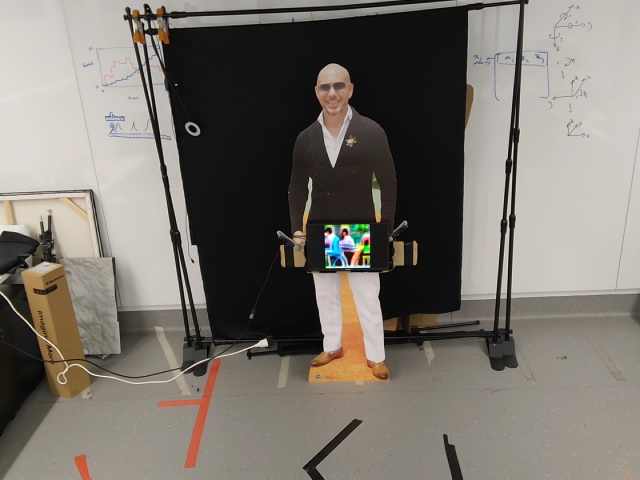}
        \caption{Top view}
        \label{fig:top}
    \end{subfigure}
    \caption{Demonstration of physical evaluation from multi-view perspectives.}
    \label{fig:views}
\end{figure}

In our multi-angle analysis, we expect patch performance to follow a V pattern. The strongest attacks are expected to occur when the patch and cut-out are angled straight on with the camera, with attack performance worsening as viewing angles become more extreme. This is roughly true and not consistently so. It is remarkable that at extreme angles, adversarial performance can still be maintained, despite the warping of the patches and reduced surface area, as evident in \Cref{fig:views}. This phenomenon indicates a level of robustness in the training process to geometric variations. All in all, the superpixel patches exhibit a stronger adversarial effect across the viewing angles considered.



\subsection{Physical evaluation using real people}\label{sec:physical-people}

\begin{figure*}[ht]
    \centering
    \begin{subfigure}{\linewidth}
        \centering
        \includegraphics[width=\linewidth]{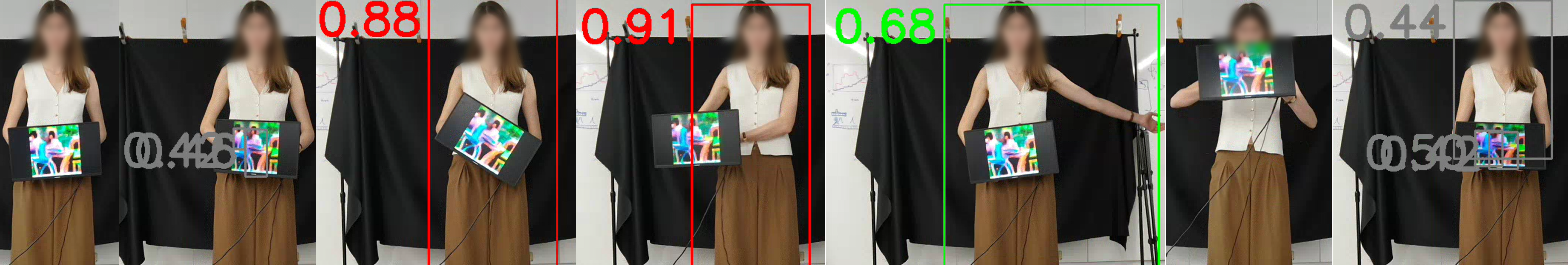}
        \caption{SPAP-1}
        \label{fig:physical_cluster}
    \end{subfigure}

    \begin{subfigure}{\linewidth}
        \centering
        \includegraphics[width=\linewidth]{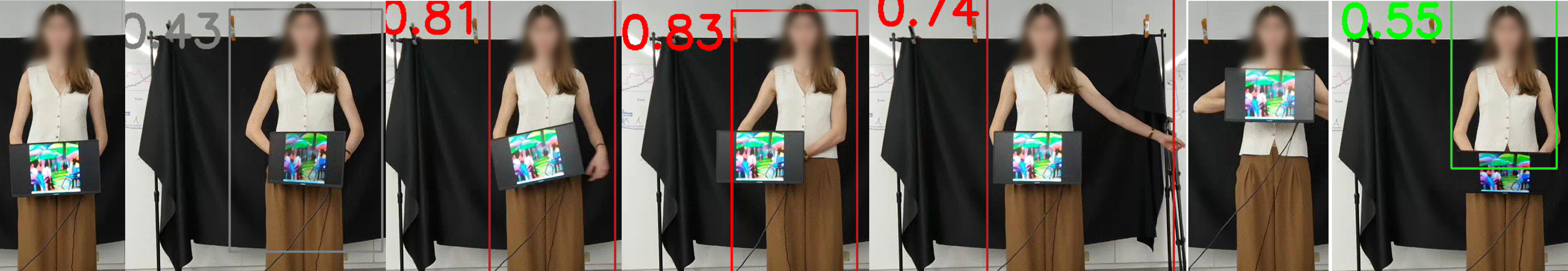}
        \caption{AdvPatch \cite{thys2019fooling}}
        \label{fig:physical_advpatch}
    \end{subfigure}
    \captionsetup{margin=0pt,skip=0pt}
    \caption{Testing patches on challenging cases. We show the patches fail to fool person detection under most translation and rotation cases.}
    \label{fig:combined_comparison}
\end{figure*}

We compare SPAP-1 to AdvPatch \cite{thys2019fooling} in a range of challenging scenarios displayed in \Cref{fig:physical_cluster} and \Cref{fig:physical_advpatch}. 
Since each frame in \Cref{fig:physical_cluster} is only an approximate replica of the corresponding frame in \Cref{fig:physical_advpatch}, the objectness scores for the two corresponding frames cannot be directly compared.
We show that under a range of rotations, translations and extension of body movements, both types of patches demonstrate partial success in suppressing detection. We highlight that all `well-performing' adversarial patches are still sensitive to lighting, the subject's pose as well as the subject's position and orientation relative to the camera.

\section{Discussion}
Experimental results in the digital domain, as reported in \Cref{sec:digital}, confirm the improvement in attack efficacy provided by our superpixel regularization technique over existing attacks. Training patches with superpixel constraints permit scale-resilience, enabling our patches to remain more adversarial under a greater number of test cases than other attacks. This performance gain is verifiably preserved in the physical world across most of the real-world scenarios tested, leveraging the systematic and controlled approach to physical evaluation reported in \Cref{sec:physical-cutout,sec:physical-people}, which is an improvement to previous studies.

Our findings suggest that constraining patch training with spatially structured features has a strong adversarial influence. Typical pipelines only consider training in the pixel or color space. Our approach opens a promising avenue for further exploration into optimization of coarser, more robust features to enhance adversarial performance.

Digital-domain experiments in the black-box setting, as reported in \Cref{sec:black-box}, demonstrate strong transferability of our attacks to some black-box models. The black-box experiments were conducted in the physical domain, but the transferability results were not good enough to be reported. This lack of transferability in the physical world is exacerbated by the degradations that occur when moving from the digital to the physical world, highlighting the need for further transfer-based regularization techniques to be developed --- a crucial, but tangential problem to the quest for scale-resilience.

\subsection{Ethical considerations}
We acknowledge that PAAs present a potential risk to the security of machine learning systems. We emphasize that the aim of this research is not to promote malicious use, but instead to highlight and explore the vulnerabilities that are inherent in deep learning models, through varying assumptions of the attacker's threat model. Our goal is to systematically evaluate these attacks in realistic settings, to help guide the research towards learning about these vulnerabilities, in order to improve the robustness of models to such attacks.

\section{Conclusion and future work}
In this study we propose a novel PAA regularizer that utilizes dynamically optimized superpixels to improve the scale-resilence of adversarial patches.
A significant contribution is deriving the gradients associated with SLIC using the Implicit Function Theorem and backpropagating the gradients in the patch training pipeline to enable intelligent clustering of superpixels. To address the lack of objective and repeatable evaluation for PAAs, we propose a protocol to allow fair comparison of patches for person detection and provide concrete evidence of the strong real-world robustness of our PAA.

Our results indicate that coarse, spatial structure-based attack optimization is a promising avenue for further exploration. 
The inclusion of further regularization techniques is expected to broaden the range of cases our attacks is effective over.
For example, introducing transferability regularization techniques such as training with surrogate models would improve attack efficacy in the black-box setting~\cite{huang2023t-sea}; a more practical threat model in the physical world.
Further regularization for real-world scenarios are an avenue to explore to reduce the visual degradation of patches that occurs naturally when they are presented in the physical world. 
Moreover, the current work can be extended by considering deformable media for the patches to be fabricated on. Clothing-based PAAs~\cite{xu2020adversarial, hu_adversarial_2022, hu_physically_2023} could improve attack efficacy and stealth. Thus, designing clothing-based PAA with our superpixel clustering method would be interesting to explore.


\bibliographystyle{IEEEtran}
\bibliography{main}

\end{document}